\documentclass[10pt,twocolumn,letterpaper]{article}

\usepackage{cvpr}              
\usepackage{graphicx}
\usepackage{balance}
\usepackage{booktabs} 
\usepackage[dvipsnames]{xcolor}

\usepackage{pifont}

\newcommand{\cmark}{\textcolor{GreenColor}{\ding{51}}\xspace}

\usepackage[pagebackref,breaklinks,colorlinks]{hyperref}
\usepackage{float}
\usepackage{amsmath}
\newcommand{\qheading}[1]{\noindent\mbox{\textbf{#1}}}
\newcommand{\RNum}[1]{\uppercase\expandafter{\romannumeral #1\relax}}
\definecolor{DeltaColor}{rgb}{0.039,0.73,0.71}
\definecolor{SigmaColor}{rgb}{0.98,0.45,0.0}
\definecolor{AlphaColor}{rgb}{0,0,0.8}
\definecolor{BetaColor}{rgb}{0.8,0,0.8}
\definecolor{GammaColor}{rgb}{0.514,0.34,0.224}
\definecolor{EpsilonColor}{rgb}{0.353,0.725,0.906}
\definecolor{GreenColor}{rgb}{0.137,0.573,0.565}
\definecolor{RedColor}{rgb}{0.949,0.275, 0.224}
\definecolor{BlueColor}{rgb}{0.0,0.0, 0.99}
\definecolor{citecolor}{HTML}{0071bc}

\title{Cloth2Tex: A Customized Cloth Texture Generation Pipeline for 3D Virtual Try-On}

\author{%
\textbf{Daiheng Gao}$^{1*}$ \quad \textbf{Xu Chen}$^{2,3*}$ \quad \textbf{Xindi Zhang}$^{1}$ \quad \textbf{Qi Wang}$^1$ \\
\textbf{Ke Sun}$^1$ \quad \textbf{Bang Zhang}$^1$ \quad \textbf{Liefeng Bo}$^1$ \quad \textbf{Qixing Huang}$^4$\\
$^1$Alibaba XR Lab \quad $^2$ETH Zurich, Department of Computer Science \\ $^3$Max Planck Institute for Intelligent Systems \quad $^4$The University of Texas at Austin\\
}

\begin{document}

\twocolumn[{%
\renewcommand\twocolumn[1][]{#1}%
\maketitle
\begin{center}
    \centering
    \captionsetup{type=figure}
    \includegraphics[width=1.0\textwidth]{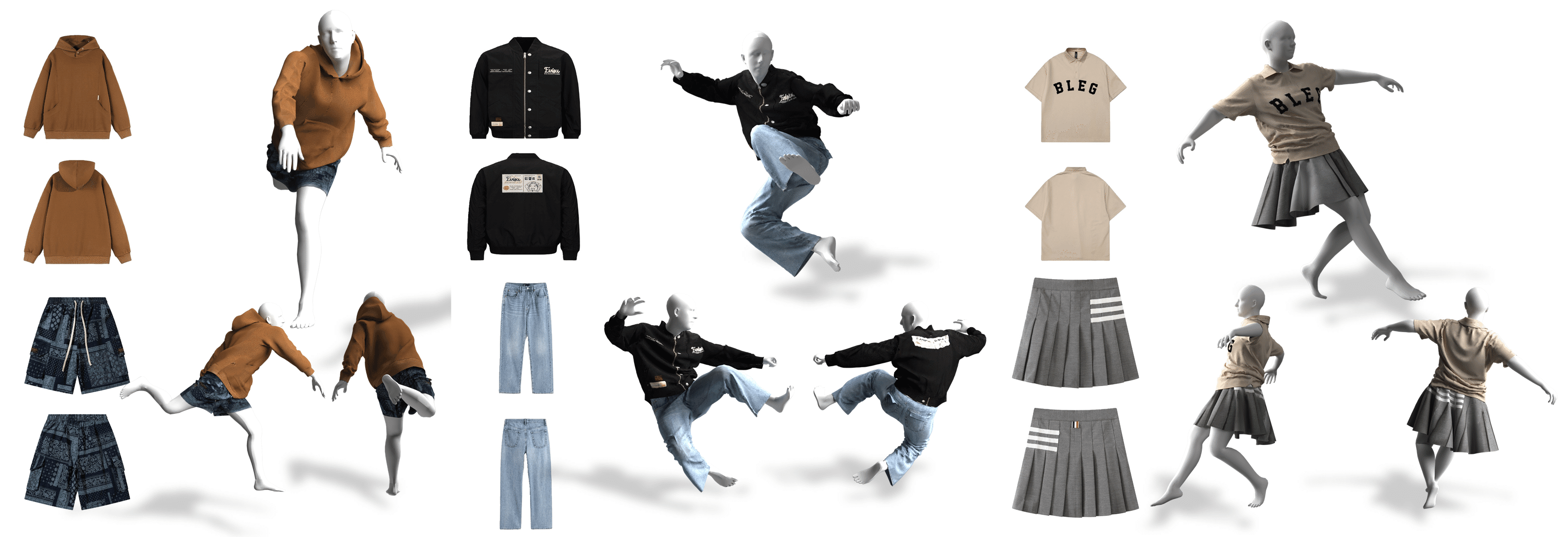}
    \captionof{figure}{We propose \textbf{Cloth2Tex}, a novel pipeline for converting 2D images of clothing to high-quality 3D textured meshes that can be draped onto 3D humans. In contrast to previous methods, Cloth2Tex supports a variety of clothing types. Results of 3D textured meshes produced by our method as well as the corresponding input images are shown above.}
\end{center}%
}]


\newcommand{\method}{Cloth2Tex\xspace}

\newcommand{\projectURL}{\href{https://tomguluson92.github.io/projects/cloth2tex}{\tt{tomguluson92.github.io/projects/cloth2tex/}}}

\begin{abstract}
Fabricating and designing 3D garments has become extremely demanding with the
increasing need for synthesizing realistic dressed persons for a variety of applications, \eg 3D virtual try-on, digitalization of 2D clothes into 3D apparel, and cloth animation. It thus necessitates a simple and straightforward pipeline to obtain high-quality texture from simple input, such as 2D reference images. Since traditional warping-based texture generation methods require a significant number of control points to be manually selected for each type of garment, which can be a time-consuming and tedious process. We propose a novel method, called \textbf{Cloth2Tex}, which eliminates the human burden in this process. Cloth2Tex is a self-supervised method that generates texture maps with reasonable layout and structural consistency. Another key feature of Cloth2Tex is that it can be used to support high-fidelity texture inpainting. This is done by combining Cloth2Tex with a prevailing latent diffusion model. We evaluate our approach both qualitatively and quantitatively and demonstrate that Cloth2Tex can generate high-quality texture maps and achieve the best visual effects in comparison to other methods. Project page: \projectURL
\end{abstract}
\section{Introduction}
\label{sec: intro}

The advancement of AR/VR and 3D graphics has opened up new possibilities for the fashion e-commerce industry. Customers can now virtually try on clothes on their avatars in 3D, which can help them make more informed purchase decisions. However, most clothing assets are currently presented in 2D catalog images, which are incompatible with 3D graphics pipelines. Thus it is critical to produce 3D clothing assets automatically from these existing 2D images, aiming at making 3D virtual try-on accessible to everyone.

\begin{figure}[t]
\includegraphics[width=1.0\linewidth]{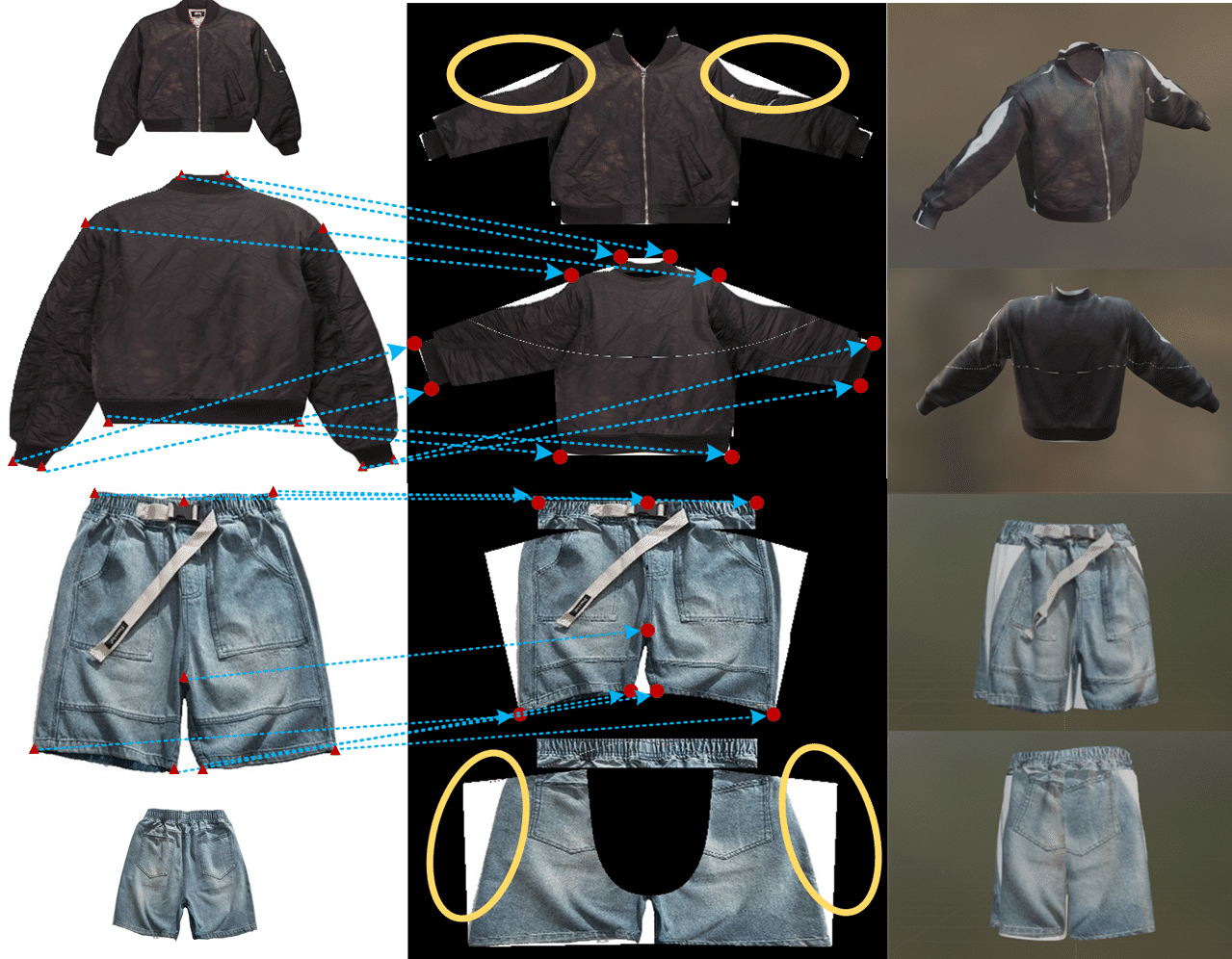}
  \caption{Problem of warping-based texture generation algorithm: partially filled UV texture maps with large missing holes as highlighted in \textcolor{yellow}{yellow}.}
\label{fig:warp_prob}
\vspace{-0.2in}
\end{figure}


Towards this goal, the research community has been developing algorithms~\cite{xu20193d,pix2surf,majithia2022robust} that can transfer 2D images into 3D textures of clothing mesh models. The key to producing 3D textures from 2D images is to determine the correspondences between the catalog images and the UV textures. Conventionally, this is achieved via the Thin-Plate-Spline (TPS) method~\cite{berg2006shape}, which approximates the dense correspondences from a small set of corresponding key points. In industrial applications, these key points are annotated manually and densely for each clothing instance to achieve good quality. With deep learning models, automatic key point detectors~\cite{brcnn, majithia2022robust} have been proposed to detect key points automatically for clothing. However, as seen in \cref{fig:warp_prob}, the inherent self-occlusions (\eg sleeves occluded by the main fabric) of TPS warping-based approaches are intractable, leading to erroneous and incomplete texture maps. Several works have attempted to use generative models to refine texture maps. However, such a refinement strategy has demonstrated success only in a small set of clothing types, \ie T-shirts, pants, and shorts. This is because TPS cannot produce satisfactory initial texture maps on all clothing types, and a large training dataset covering high-quality texture maps of diverse clothing types is missing. Pix2Surf~\cite{pix2surf}, a SMPL~\cite{loper2015smpl}-based virtual try-on algorithm, has automated the process of texture generation with no apparent cavity or void. However, due to its clothing-specific model, Pix2Surf is limited in its ability to generalize to clothes with arbitrary shapes.

This paper aims to automatically convert 2D reference clothing images into 3D textured clothing meshes for a larger diversity of clothing types. To this end, we first contribute template mesh models for 10+ different clothing types (well beyond current SOTAs: Pix2Surf (\textbf{4}) and ~\cite{majithia2022robust} (\textbf{2})). Next, instead of using the Thin-Plate-Spline (TPS) method as previous methods, we incorporate neural mesh rendering~\cite{softras} to directly establish dense correspondences between 2D catalog images and the UV textures of the meshes. This results in higher-quality initial texture maps for all clothing types. We achieve this by optimizing the 3D clothing mesh models and textures to align with the catalog images' color, silhouette, and key points.

Although the texture maps from neural rendering are of higher quality, they still need refinement due to missing regions. Learning to refine these texture maps across different clothing types requires a large dataset of high-quality 3D textures, which is infeasible to acquire. We tackle this problem by leveraging the recently emerging latent diffusion model (LDM)~\cite{stablediffusion} as a data simulator. Specifically, we use ControlNet~\cite{controlnet} to generate large-scale, high-quality texture maps with various patterns and colors based on its \textit{canny edge} version. In addition to the high-quality ground-truth textures, the refinement network requires the corresponding initial defective texture maps obtained from neural rendering. To get such data, we render the high-quality texture maps into catalog images and then run our neural rendering pipeline to re-obtain the texture map from the catalog images, which now contain defects as desired. With these pairs of high-quality complete texture maps and the defective texture maps from the neural renderer, we train a high-resolution image translation model that refines the defective texture maps.

Our method can produce high-quality 3D textured clothing from 2D catalog images of various clothing types. In our experiments, we compare our approach with state-of-the-art techniques of inferring 3D clothing textures and find that our method supports more clothing types and demonstrates superior texture quality. In addition, we carefully verify the effectiveness of individual components via a thorough ablation study.

In summary, we contribute \textbf{Cloth2Tex}, a pipeline that can produce high-quality 3D textured clothing in various types based on 2D catalog images, which is achieved via
\begin{itemize}
    \item \textit{a)} 3D parametric clothing mesh models of 10+ different categories that will be publicly available,
    \item \textit{b)} an approach based on neural mesh rendering to transferring 2D catalog images into texture maps of clothing meshes,
    \item \textit{c)} data simulation approach for training a texture refinement network built on top of blendshape-driven mesh and LDM-based texture.
\end{itemize}
\section{Related Works}

\qheading{Learning 3D Textures.} Our method is related to learning texture maps for 3D meshes.
Texturify~\cite{siddiqui2022texturify} learns to generate high-fidelity texture maps by rendering multiple 2D images from different viewpoints and aligning the distribution of rendered images and real image observations. Yu ~\etal~\cite{yu2021learning} adopt a similar method, rendering images from different viewpoints and then discriminating the images by separate discriminators. With the emergence of diffusion models~\cite{dhariwal2021diffusion, trabucco2023effective}, recent work Text2Tex~\cite{chen2023text2tex} exploits 2D diffusion models for 3D texture synthesis. Due to the mighty generalization ability of the diffusion model~\cite{ho2020denoising,stablediffusion} trained on the largest corpus LAION-5B~\cite{laion5b}, \ie stable diffusion~\cite{stablediffusion}, the textured meshes generated by Text2Tex are of superior quality and contain rich details. Our method is related to these approaches in that we also utilize diffusion models for 3D texture learning. However, different from previous approaches, we use latent diffusion models only to generate synthetic texture maps to train our texture inpainting model, and our focus lies in learning 3D texture corresponding to a specific pair of 2D reference images instead of random or text-guided generation.

\qheading{Texture-based 3D Virtual Try-On.} Wang \etal~\cite{garmentdesign_Wang_SA18} provide a sketch-based network that infers both 2D garment sewing patterns and the draped 3D garment mesh from 2D sketches. 
In real applications, however, many applications require inferring 3D garments and the texture from 2D catalog images.
To achieve this goal, Pix2Surf~\cite{pix2surf} is the first work that creates textured 3D garments automatically from front/back view images of a garment. This is achieved by predicting dense correspondences between the 2D images and the 3D mesh template using a trained network. However, due to the erroneous correspondence prediction, particularly on unseen test samples, Pix2Surf has difficulty in preserving high-frequency details and tends to blur out fine-grained details such as thin lines and logos.


To avoid such a problem, Sahib \etal~\cite{majithia2022robust} propose to use a warping-based method (TPS)~\cite{berg2006shape} instead and to use further a deep texture inpainting network built upon MADFNet~\cite{madf}. However, as mentioned in the introduction, warping-based methods generally require dense and accurate corresponding key points in images and UV maps and have only demonstrated successful results on two simple clothing categories, T-shirts and trousers. In contrast to previous work, \method aims to achieve automatic high-quality texture learning for a broader range of garment categories. To this end, we use neural rendering instead of warping, which yields better texture quality on more complex garment categories. We further utilize latent diffusion models (LDMs) to synthesize high-quality texture maps of various clothing categories to train the inpainting network.


\begin{figure*}[t]
\includegraphics[width=1\linewidth]{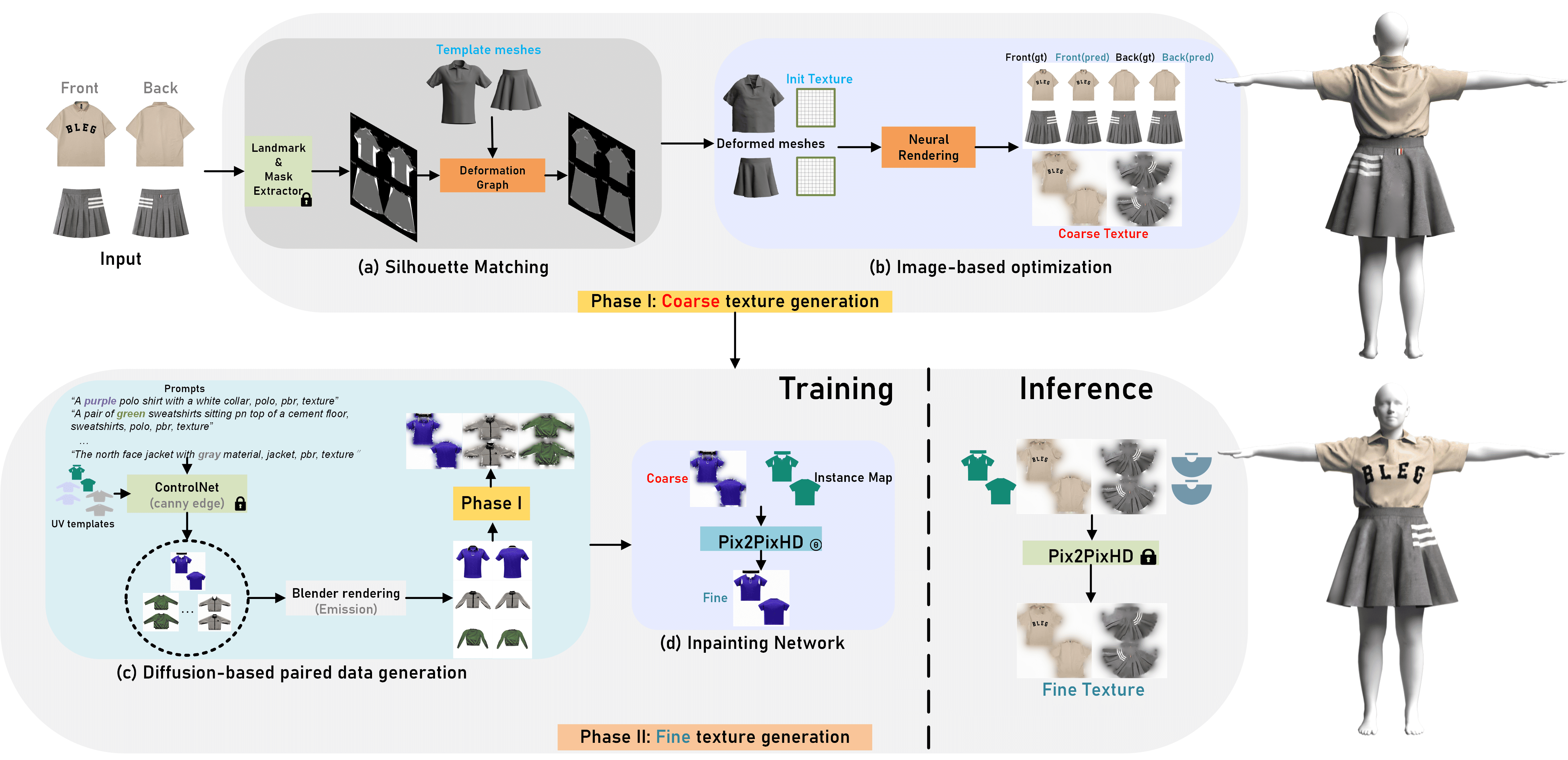}
  \caption{\textbf{Method overview}: \method consists of two stages. In Phase \RNum{1}, we determine the 3D garment shape and coarse texture by registering our parametric garment meshes onto catalog images using a neural mesh renderer. Next, in Phase \RNum{2}, we refine the coarse estimate of the texture to obtain high-quality fine textures using image translation networks trained on large-scale data synthesized by pre-trained latent diffusion models. Note that the only component that requires training is the inpainting network. Please watch our video on the project page for an animated explanation of \method.}
\label{fig:maingraph}
\vspace{-0.2in}
\end{figure*}

\section{Method}
We propose \method, a two-stage approach that converts 2D images into textured 3D garments. The garments are represented as polygon meshes, which can be draped and simulated on 3D human bodies. The overall pipeline is illustrated in \cref{fig:maingraph}. The pipeline's first stage (Phase \RNum{1}) is to determine the 3D garment shape and coarse texture. We do this by registering our parametric garment meshes onto catalog images using a neural mesh renderer. The pipeline's second stage (Phase \RNum{2}) is to recover fine textures from the coarse estimate. We use image translation networks trained on large-scale data synthesized by pre-trained latent diffusion models. The mesh templates for individual clothing categories are a pre-requirement for our pipeline. We obtain these templates by manual artist design and will make them publicly available. 

Implementation details are placed in the supp. material due to the page limit.

\subsection{Pre-requirement: Template Meshes}
\label{sec:pre}

For the sake of both practicality and convenience, we design cloth template mesh (with fixed topology) $\mathcal{M}$ for common garment types (\eg, T-shirts, sweatshirts, baseball jackets, trousers, shorts, skirts, and \etc). We then build a deformation graph $\mathcal{D}$ ~\cite{deformgraph} to optimize the template mesh vertices. This is because per-vertex image-based optimization is subject to errors and artifacts due to the high degrees of freedom. Specifically, we construct $\mathcal{D}$ with $k$ nodes, which are parameterized with axis angles $\mathbf{A} \in \mathbb{R}^{3}$ and translations $\mathbf{T} \in \mathbb{R}^{3}$. The vertex displacements are then derived from the deformation nodes (the number of nodes $k$ is dependent on the garment type since different templates have different numbers of vertices and faces). We also manually select several vertices on the mesh templates as landmarks $\mathcal{K}$. The specific requirements of the template mesh are as follows: vertices $V$ less than 10,000, uniform mesh topology, and integrity of UV. The vertex number of all templates ranges between \textbf{skirt} (\textit{6,116}) to \textbf{windbreaker} (\textit{9,881}). For uniformity, we set the downsampling factor of $\mathcal{D}$ for all templates to 20 (details of template meshes are placed in the supp. material). The integrity of UV means that the UV should be placed as a whole in terms of front and back, without further subdivision, as used in traditional computer graphics. Fabricating integral UV is not complicated and can be a super-duper candidate for later diffusion-based texture generation. See \cref{sec:phase3} for more details.

\subsection{Phase \RNum{1}: Shape and Coarse Texture Generation}
\label{sec:phase1}

The goal of Phase \RNum{1} is to determine the garment shape and a coarse estimate of the UV textures $\mathcal{T}$ from the input catalog (\textit{Front \& Back} view). We adopt a differentiable rendering approach~\cite{softras} to determine the UV textures in a self-supervised way without involving trained neural networks. Precisely, we fit our template model to the catalog images by minimizing the difference between the 2D rendering of our mesh model and the target images. The fitting procedure consists of two stages, namely \textit{Silhouette Matching} and \textit{Image-based Optimization}. We will now elaborate on these stages below.

\subsubsection{Silhouette Matching}

We first align the corresponding template mesh to the 2D images based on the 2D landmarks and silhouette. Here, we use BCRNN~\cite{brcnn} to detect landmarks $L_{2d}$ and DenseCLIP~\cite{denseclip} to extract the silhouette $M$. To fit our various types of garments, we finetune BCRNN with 2,000+ manually annotated clothing images per type.

After the mask and landmarks of the input images are obtained, we first perform a global rigid alignment by an automatic cloth scaling method to adjust the scaling factor of mesh vertices according to the overlap of the initial silhouette between mesh and input images, which ensures a rough agreement of the yielded texture map (See \cref{fig:ablation_phase1}). Specifically, we implement this mechanism by checking the silhouette between the rendered and reference images, and then enlarging or shrinking the scale of mesh vertices accordingly. After an optimum \textbf{Intersection over Union(IoU)} has been achieved, we fix the coefficient and send the scaled template to the next step.

We then fit the silhouette and the landmarks of the template mesh (the landmarks on the template mesh are pre-defined as described in \cref{sec:pre}) to those detected from the 2D catalog images. To this end, we optimize the deformations of the nodes in the deformation graph by minimizing the following energy terms:

\qheading{2D Landmark Alignment $E_{\textup{lmk}}$}
measures the distance between 2D landmarks $L_{\textup{2d}}$ detected by BRCNN and the 2D projection of 3D template mesh keypoints:

 \begin{align}
    \label{eq:landmark2d}
    E_{\textup{lmk}}= \lVert \prod\mathcal{K} - L_{\textup{2d}}\rVert_2
\end{align}
where $\prod$ denotes the 2D projection of 3D keypoints.

\qheading{2D Silhouette Alignment $E_{\textup{sil}}$}
measures the overlap between the silhouette of $\mathcal{M}$ and the predicted $M$ from DenseCLIP:
\begin{align}
\label{eq:mask}
E_{\textup{sil}}= \textup{MaskIoU}(S_{\textup{proj}}(\mathcal{M}), M)
\end{align}
where $S_{\textup{proj}}(\mathcal{M})$ is the silhouette rendered by the differentiable mesh renderer SoftRas~\cite{softras} and \textit{MaskIoU} loss is derived from Kaolin~\cite{KaolinLibrary}.

Merely minimizing $E_{\textup{lmk}}$ and  $E_{\textup{sil}}$ does not lead to satisfactory results, and optimization procedure can easily get trapped into local minimums. To alleviate this issue, we introduce a couple of regularization terms. We first regularize the deformation using the as-rigid-as-possible loss $E_{\textup{arap}}$ ~\cite{sorkine2007rigid} which penalizes the deviation of estimated local surface deformations from rigid transformations. Moreover, we further enforce the normal consistency $E_{\textup{norm}}$, which measures normal consistency for each pair of neighboring faces). The overall optimization objective is given as:
\begin{align}
\label{eq:sh}
w_{\textup{sil}}E_{\textup{sil}} + w_{\textup{lmk}}E_{\textup{lmk}} + w_{\textup{arap}}E_{\textup{arap}} + w_{\textup{norm}}E_{\textup{norm}}
\end{align}
where $w_*$ are the respective weights of the losses.

We set large regularization weights $w_{\textup{arap}}$, $w_{\textup{norm}}$ at the initial iterations. We then reduce their values progressively during the optimization procedure, so that the final rendered texture aligns with the input images. Please refer to the supp. material for more details. 

\subsubsection{Image-based Optimization}

After the shape of the template mesh is aligned with the image silhouette, we then optimize the UV texture map $\mathcal{T}$ to minimize the difference between the rendered image $I_{\textup{rend}}=S_{\textup{rend}}(\mathcal{M},\mathcal{T})$ and the given input catalog images $I_{\textup{in}}$ from both sides simultaneously. To avoid any outside interference during the optimization, we only preserve the ambient color and set both diffuse and specular components to be zero in the settings of SoftRas~\cite{softras}, PyTorch3D~\cite{ravi2020pytorch3d}.




Since the front and back views do not cover the full clothing texture, \eg the seams between the front and back bodice can not be recovered well due to the occlusions, 
we use the total variation method~\cite{rudin1992nonlinear} to fill in the blank of seam-affected UV areas. The total variation loss $E_{\textup{tv}}$ is defined as the norm of the spatial gradients of the rendered image $\nabla_x I_{\textup{rend}}$ and $\nabla_y I_{\textup{rend}}$:
\begin{equation}
\begin{aligned}
    \label{eq:tv}
    E_{tv}= \lVert \nabla_x I_{\textup{rend}} \rVert_2  + \lVert \nabla_y I_{\textup{rend}} \rVert_2
\end{aligned}
\end{equation}

In summary, the energy function for the image-based optimization is defined as below:
\begin{align}
    \label{eq:image_opt}
   w_{\textup{img}}\lVert I_{\textup{in}} - I_{\textup{rend}} \rVert_2 + w_{\textup{tv}}E_{\textup{tv}}
\end{align}
where $I_{\textup{in}}$ and $I_{\textup{rend}}$ are the reference and rendered image. As shown in \cref{fig:maingraph}, $\mathcal{T}$ implicitly changes towards the final coarse texture $\mathcal{T}_{coarse}$, which ensures the final rendering is as similar as possible with the input. Please refer to our attached video for a vivid illustration.

\subsection{\textbf{Phase \RNum{2}: Fine texture generation}}

In Phase \RNum{2}, we refine the coarse texture from \cref{sec:phase1} and fill in the missing regions. Our approach takes inspiration from the strong and comprehensive capacity of Stable Diffusion (SD), which is a terrific model to have by itself in image inpainting, completion, and text2image tasks. In fact, there's also an entire, growing ecosystem around it: LoRA~\cite{hu2021lora}, ControlNet~\cite{controlnet}, textual inversion~\cite{gal2022text_inversion} and Stable Diffusion WebUI~\cite{SDWebUI}. Therefore, a straightforward idea is to resolve our texture completion via SD. 

However, we find poor content consistency between the inpainted blank and original textured UV. This is because UV data in our setting rarely appears in the training dataset LAION-5B~\cite{laion5b} of SD. In other words, the semantic composition of LAION-5B and UV texture (cloth) are quite different and challenging for SD to generalize.

To address this issue, we first leverage ControlNet~\cite{controlnet} to generate $\sim 2,000+$ HQ complete textures per template and render emission-only images under the front and back view. Next, we use Phase \RNum{1} again to recover the corresponding coarse textures. After collecting the pairs of coarse and fine textures, we train an inpainting network to fill the missing regions in the coarse texture maps. 

\subsubsection{Diffusion-based Data Generation}
\label{sec:phase3}

We employ diffusion models~\cite{stablediffusion,controlnet, dhariwal2021diffusion} to generate realistic and diverse training data. 

We generate texture maps following the UV template configuration, adopting the pre-trained ControlNet with edge map as input conditions. ControlNet finetunes text-to-image diffusion models to incorporate additional structural conditions as input. The input edge maps are obtained through canny edge detection on clothing-specific UV, and the input text prompts are generated by applying image captioning models, namely Lavis-BLIP~\cite{lavis}, OFA~\cite{ofa} and MPlug~\cite{mplug}, on tens of thousands of clothes crawled from Amazon and Taobao.

\begin{figure*}[htpb]
\includegraphics[width=1.0\linewidth]{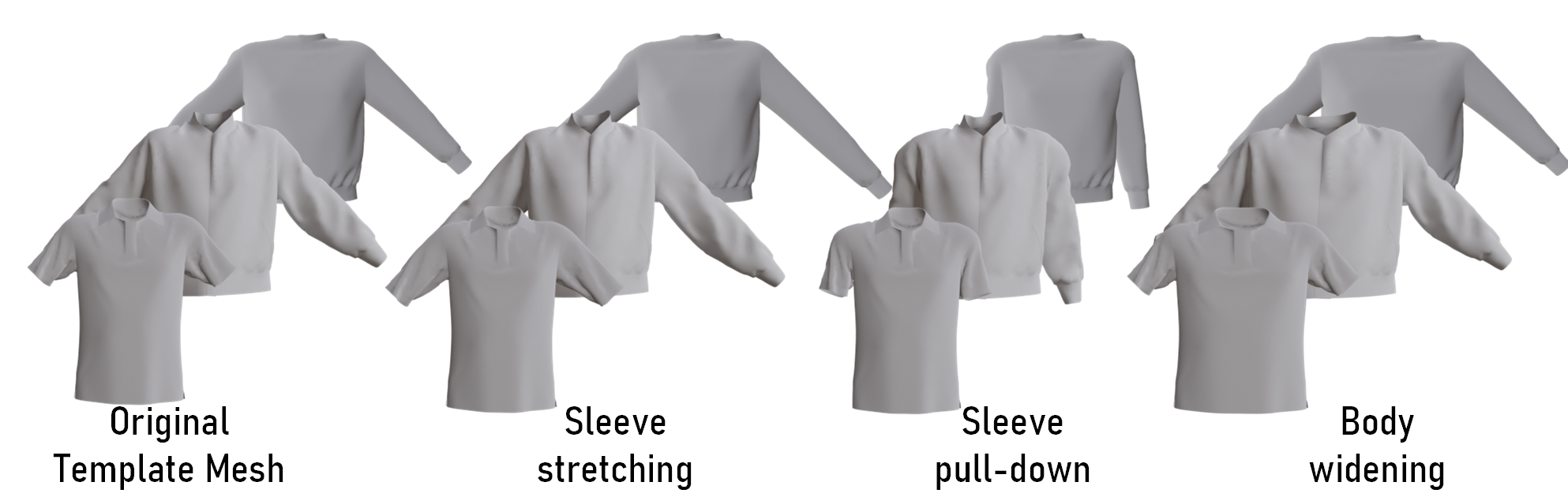}
  \caption{Illustration of the three sleeve-related blendshapes of our template mesh model. These blendshapes allow rendering clothing images in diverse pose configurations to facilitate simulating real-world clothing image layouts.}
\label{fig:blendshape}
\vspace{-0.2in}
\end{figure*}

After generating the fine UV texture maps, we are already able to generate synthetic front and back 2D catalog images, which will be used to train the impainting network. We leverage the rendering power of Blender native EEVEE engine to get the best visual result. A critical step of our approach is to perform data augmentation so that the impainting network captures invariant features instead of details that differ between synthetic images and testing images, which do not generalize. To this end, we vary the blend shape parameters of the template mesh to generate 2D catalog images in different shapes and pose configurations and simulate self-occlusions, which frequently exist in reality and lead to erroneous textures as shown in \cref{fig:warp_prob}. We hand-craft three common blendshapes (\cref{fig:blendshape}) that are enough to simulate the diverse cloth-sleeve correlation/layout in reality.


Next, we run Phase \RNum{1} to produce coarse textures from the rendered synthetic 2D catalog images, yielding the coarse, defect textures corresponding to the fine textures. These pairs of coarse-fine textures serve as the training data for the subsequent inpainting network.

\subsubsection{Texture Inpainting}
\label{sec:phase4}

Given the training data simulated by LDMs, we then train our inpainting network. Note that we train a single network for all clothing categories, making it general-purpose. 

Regarding the impainting work, we choose Pix2PixHD~\cite{pix2pixhd}, which shows better results than alternative approaches such as conditional TransUNet~\cite{chen2021transunet}, ControlNet. One issue of Pix2PixHD is that produces color-consistent output $\mathcal{T}_{o}$ in contrast to prompt-guided ControlNet (please check our supp. material for visualization comparison). These results are compared with the input full UV as the condition. To address this issue, we first locate the missing holes, continuous edges and lines in the coarse UV as the residual mask $M_{r}$ (left corner at the  bottom line of \cref{fig:qual_2}). We then linearly blend those blank areas with the model's output during texture repairing. Formally speaking, we compute the output as below:
\begin{align}
    \label{eq:final_tex}
    &\mathcal{T}_{\textup{fine}} = \textup{BilateralFilter}(\mathcal{T}_{\textup{coarse}} + M_{r} * \mathcal{T}_{o})
\end{align}
where $\textup{BilateralFilter}$ is non-linear filter that can blur the irregular and rough seaming between $\mathcal{T}_{\textup{coarse}}$ and $\mathcal{T}_{o}$ very well while keeping edges fairly sharp. More details can be seen in our attached video.
\section{Experiments}

Our goal is to generate 3D garments from 2D catalog images. We verify the effectiveness of \method via thorough evaluation and comparison with state-of-the-art baselines. Furthermore, we conduct a detailed ablation study to demonstrate the effects of individual components.




\begin{figure}[htpb]
\includegraphics[width=1.0\linewidth]{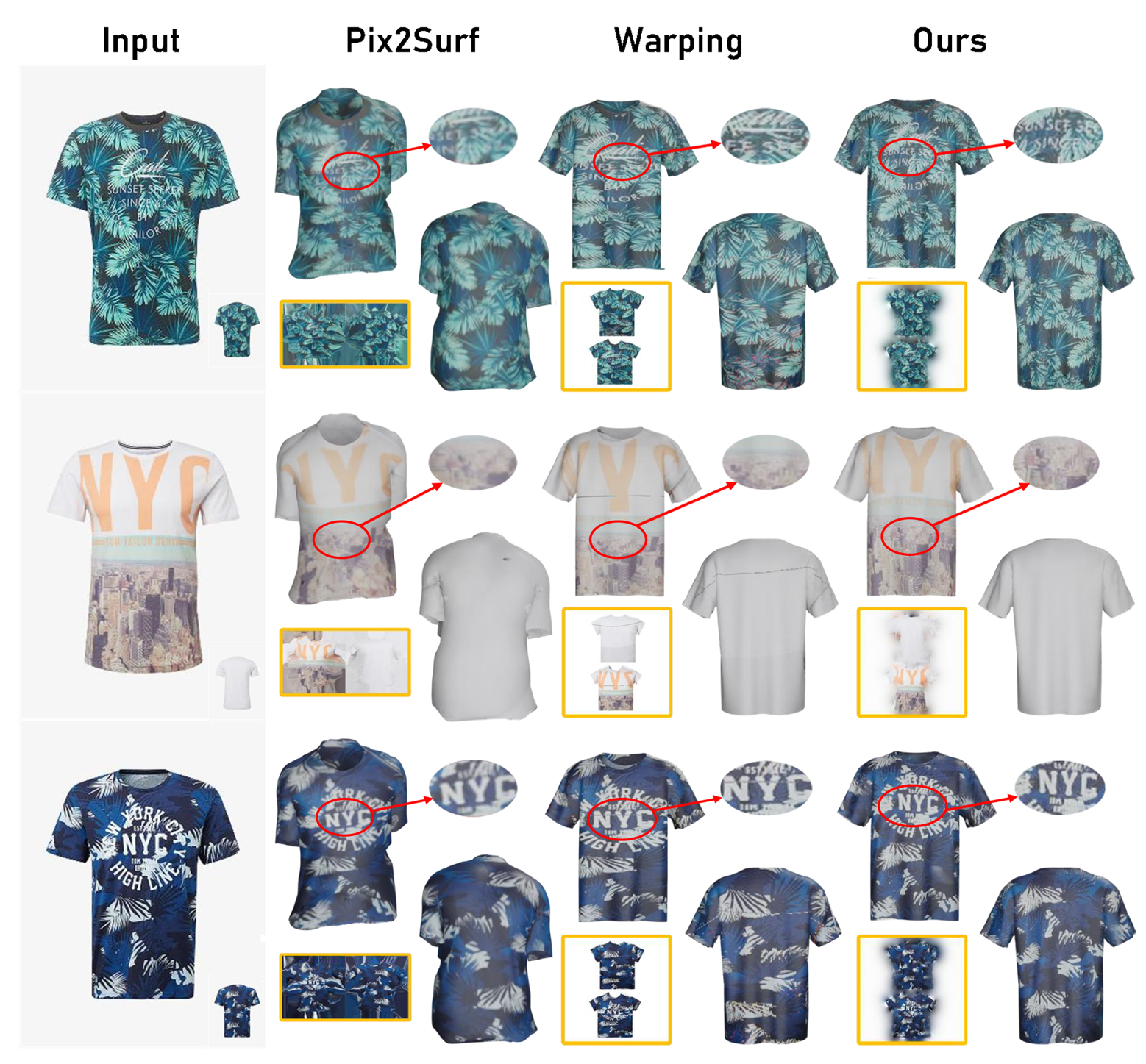}
  \caption{Comparison with Pix2Surf~\cite{pix2surf} and Warping~\cite{majithia2022robust} on T-shirts. Please zoom in for more details.}
\label{fig:qual_1}
\vspace{-0.2in}
\end{figure}

\subsection{Comparison with SOTA}

We first compare our method with SOTA virtual try-on algorithms, both 3D and 2D approaches.

\noindent \textbf{Comparison with 3D SOTA:} We compare \method with SOTA methods that produce 3D mesh textures from 2D clothing images, including model-based Pix2Surf~\cite{pix2surf} and TPS-based Warping~\cite{majithia2022robust} (We replace the original MADF with locally changed UV-constrained Naiver Stokes method, differences between our UV-constrained naiver-stokes and original version is described in the suppl. material).
As shown in \cref{fig:qual_1}, our method produces high-fidelity 3D textures with sharp, high-frequency details of the patterns on clothing, such as the leaves and characters on the top row. In addition, our method accurately preserves the spatial configuration of the garment, particularly the overall aspect ratio of the patterns and the relative locations of the logos. In contrast, the baseline method Pix2Surf~\cite{pix2surf} tends to produce blurry textures due to a smooth mapping network, and the Warping~\cite{majithia2022robust} baseline introduces undesired spatial distortions (e.g., second row in \cref{fig:qual_1}) due to sparse correspondences. 



\begin{figure}[htpb]
\includegraphics[width=1.0\linewidth]{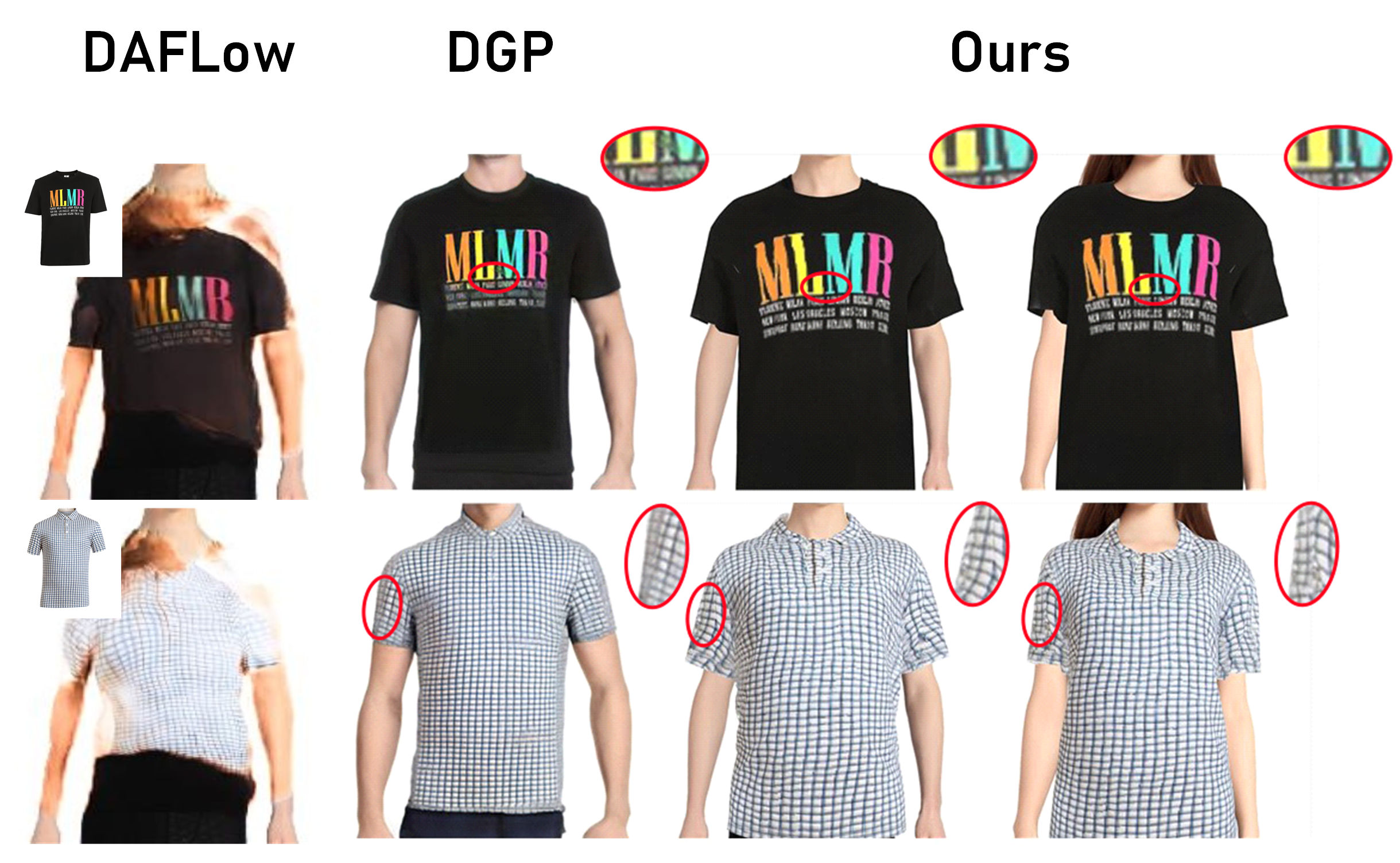}
  \caption{Comparison with 2D Virtual Try-One methods, including DAFlow~\cite{daflow_eccv22} and DGP~\cite{dgp_cvpr22}.}
\label{fig:qual_3}
\end{figure}

\noindent \textbf{Comparison with 2D SOTA:} We further compare \method with 2D virtual try-on methods: Flow-based DAFlow~\cite{daflow_eccv22} and StyleGAN-enhanced Deep-Generative-Projection (DGP)~\cite{dgp_cvpr22}.
As shown in \cref{fig:qual_3}, \method achieves better quality than 2D virtual try-on methods in sharpness and semantic consistency.
More importantly, our outputs, namely 3D textured clothing meshes, are naturally compatible with cloth physics simulation, allowing the synthesis of realistic try-on effects in various body poses. In contrast, 2D methods rely on prior learned from training images and are hence limited in their generalization ability to extreme poses outside the training distribution.






\begin{figure}[htpb]
\includegraphics[width=1.0\linewidth]{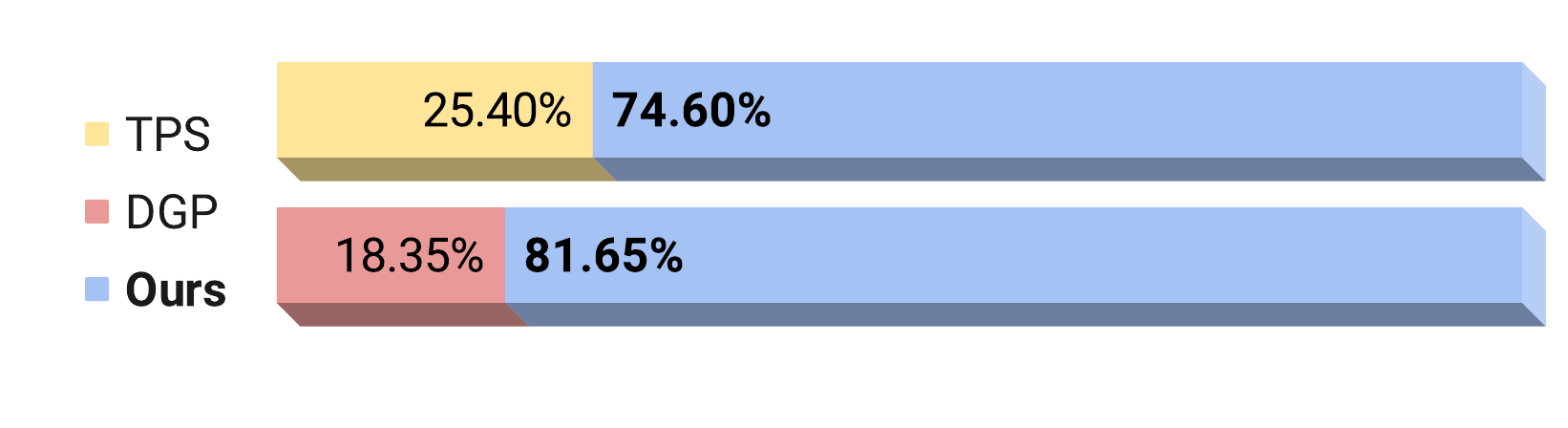}
\caption{User preferences among 643 responses from 72 participants. Our method is favored by significantly more users.}
\label{fig:user_study}
\vspace{-0.1in}
\end{figure}

\noindent \textbf{User Study:} Finally, we conduct a user study to evaluate the overall perceptual quality and consistency with our methods' provided input catalog images and 2D and 3D baselines. We consider DGP the 2D baseline and TPS the 3D baseline due to their best performance among existing work. Each participant is shown three randomly selected pairs of results, one produced by our method and the other made by one of the baseline methods. The participant is requested to choose the one that appears more realistic and matches the reference clothing image better. In total, we received 643 responses from 72 users aged between 15 and 60. The results are reported in \cref{fig:user_study}. Compared to DGP~\cite{dgp_cvpr22} and TPS, \method is favored by the participants with preference rates of 74.60\%  and 81.65\%, respectively. This user study result verified the quality and consistency of our method. 


\begin{figure}[htpb]
\includegraphics[width=1.0\linewidth]{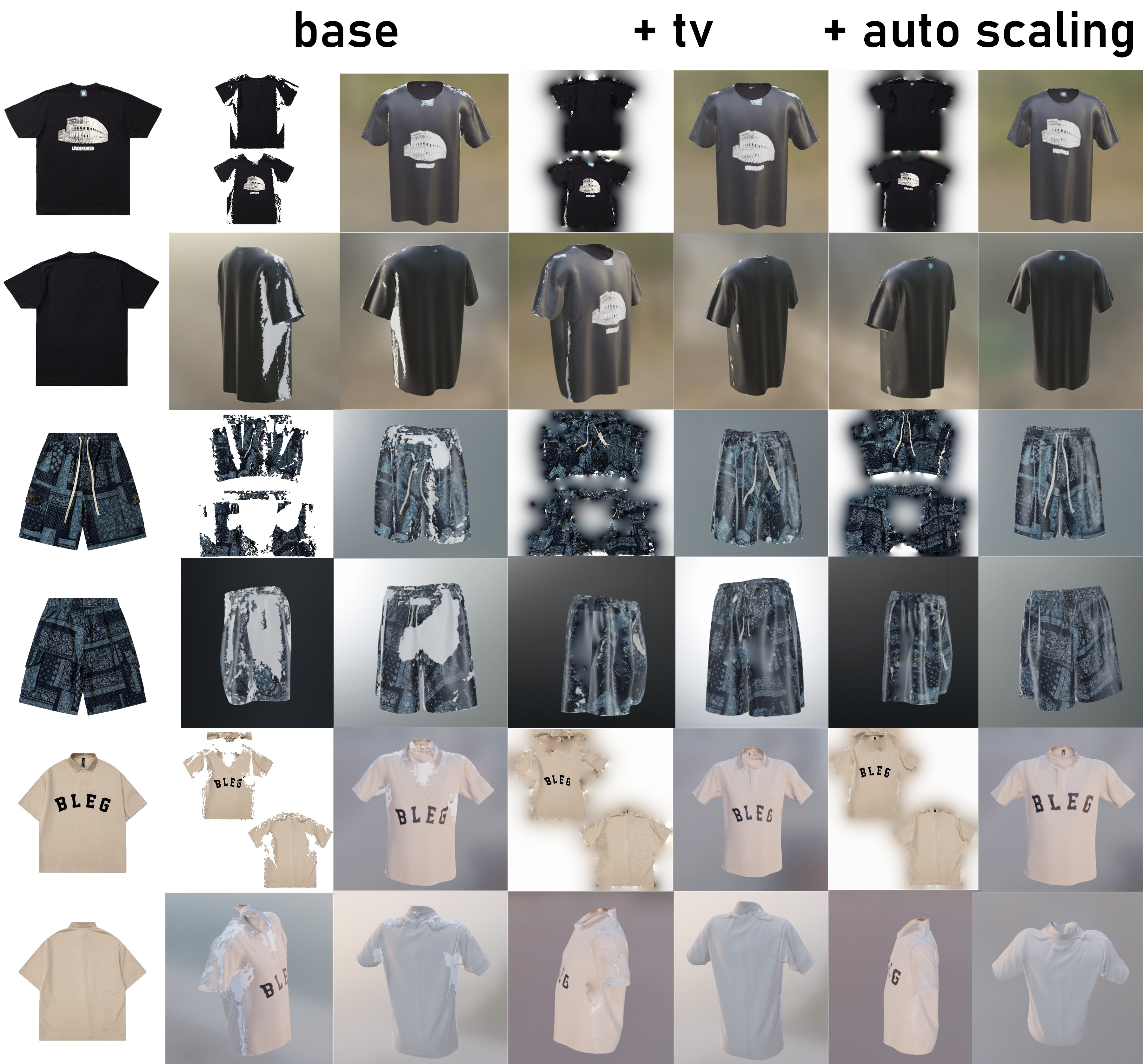}
  \caption{Ablation Study on Phase \RNum{1}. From left to right: base, base +  total variation loss $E_{\textup{tv}}$, base + $E_{\textup{tv}}$ + automatic scaling.}
\label{fig:ablation_phase1}
\end{figure}

\begin{figure*}[htpb]
\includegraphics[width=1.0\linewidth]{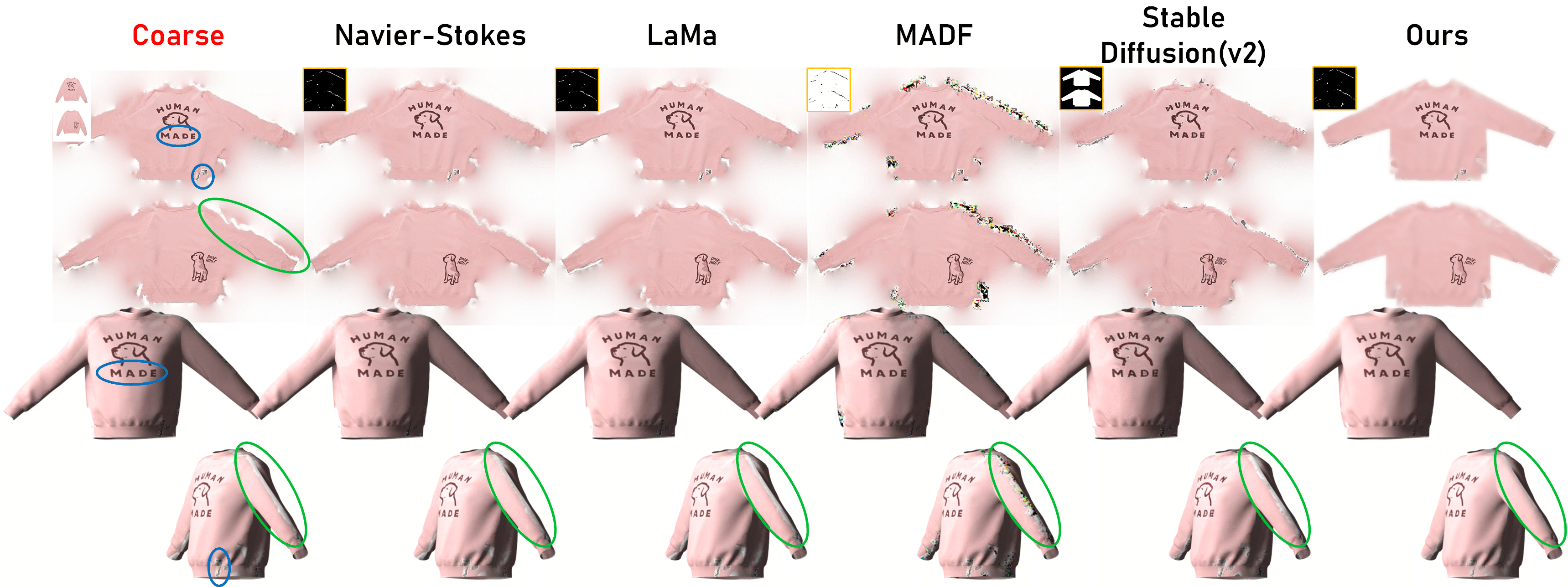}
  \caption{Comparison with SOTA inpainting methods (Naiver-Stokes~\cite{nsinpainting}, LaMa~\cite{lama}, MADF~\cite{madf} and Stable Diffusion v2~\cite{stablediffusion}) on texture inpainting. The upper left corners of each column are the conditional mask input. {\bf\color{EpsilonColor}{Blue}} in the first column shows that our method is capable of maintaining consistent boundary and curvature \textit{w.r.t} reference image while {\bf\color{GreenColor}{Green}} highlights the blank regions that need inpainting.} 
  
\label{fig:qual_2}
\end{figure*}

\subsection{Ablation Study}

To demonstrate the effect of individual components in our pipeline, we perform an ablation study for both stages in our pipeline.

\qheading{Neural Rendering vs. TPS Warping:} TPS warping has been widely used in previous work on generating 3D garment textures. However, we found that it suffers from challenging cases illustrated in \cref{fig:warp_prob}, so we propose a new pipeline based on neural rendering. We compare our method with TPS warping quantitatively to verify this design choice. 
Our test set consists of 10+ clothing categories, including T-shirts, Polos, sweatshirts, jackets, hoodies, shorts, trousers, and skirts, with 500 samples per category. 
We report the structure similarity (SSIM~\cite{ssim}) and peak signal-to-noise ratio (PSNR) between the recovered textures and the ground truth textures.

As shown in \cref{tab:ablation_phase2}, our neural rendering-based pipeline achieves superior SSIM and PSNR compared to TPS warping. This improvement is also preserved after inpainting and refinement, leading to a much better quality of the final texture. We conduct a comprehensive comparison study on various inpainting methods in the supp. material, and please check it if needed. 

\begin{table}[h]
\vspace{-0.5em}
\caption{Neural Rendering vs. TPS Warping. We evaluate the texture quality of neural rendering and TPS-based warping, with and without inpainting.}
\label{tab:ablation_phase2}
\centering
\begin{tabular}{@{}cccc@{}}
\toprule
 Baseline & Inpainting & SSIM $\uparrow$ & PSNR $\uparrow$\\
\hline
TPS  & \textit{None}  &  0.70 & 20.29    \\
TPS  & \textit{Pix2PixHD}  &  0.76 & 23.81   \\
Phase \RNum{1}  & \textit{None}  & 0.80 & 21.72  \\
Phase \RNum{1}  & \textit{Pix2PixHD}   & \textbf{0.83} & \textbf{24.56} \\
\bottomrule
\end{tabular}
\end{table}


\qheading{Total Variation Loss \& Automatic Scaling (Phase \RNum{1}) }
As shown in \cref{fig:ablation_phase1}, dropping the total variation loss $E_{tv}$ and automatic scaling, the textures are incomplete and cannot maintain a semantically correct layout. With $E_{tv}$, \method produces more complete textures by exploiting the local consistency of textures. Further applying automatic scaling results in better alignment between the template mesh and the input images, resulting in a more semantically correct texture map. 


\qheading{Inpainting Methods (Phase \RNum{2})}
Next, to demonstrate the need for training an inpainting model specifically for UV clothing textures, we compare our task-specific inpainting model with general-purpose inpainting algorithms, including Navier-Stokes~\cite{nsinpainting} algorithm and off-the-shelf deep learning models including LaMa~\cite{lama}, MADF~\cite{madf} and Stable Diffusion v2~\cite{stablediffusion} with pre-trained checkpoints. 
Here, we modify the traditional Navier-Stokes~\cite{nsinpainting} algorithm to a UV-constrained version because a texture map is only part of the whole squared image grid, where plenty of non-UV regions produce an adverse effect for texture in-painting (please see supp. material for comparison).

As shown in \cref{fig:qual_2}, our method, trained on our synthetic dataset generated by the diffusion model, outperforms general-purpose inpainting methods in the task of refining and completing clothing textures, especially in terms of the color consistency between inpainted regions and the original image. 


\begin{figure*}[htpb]
\centering
\includegraphics[width=1.0\linewidth]{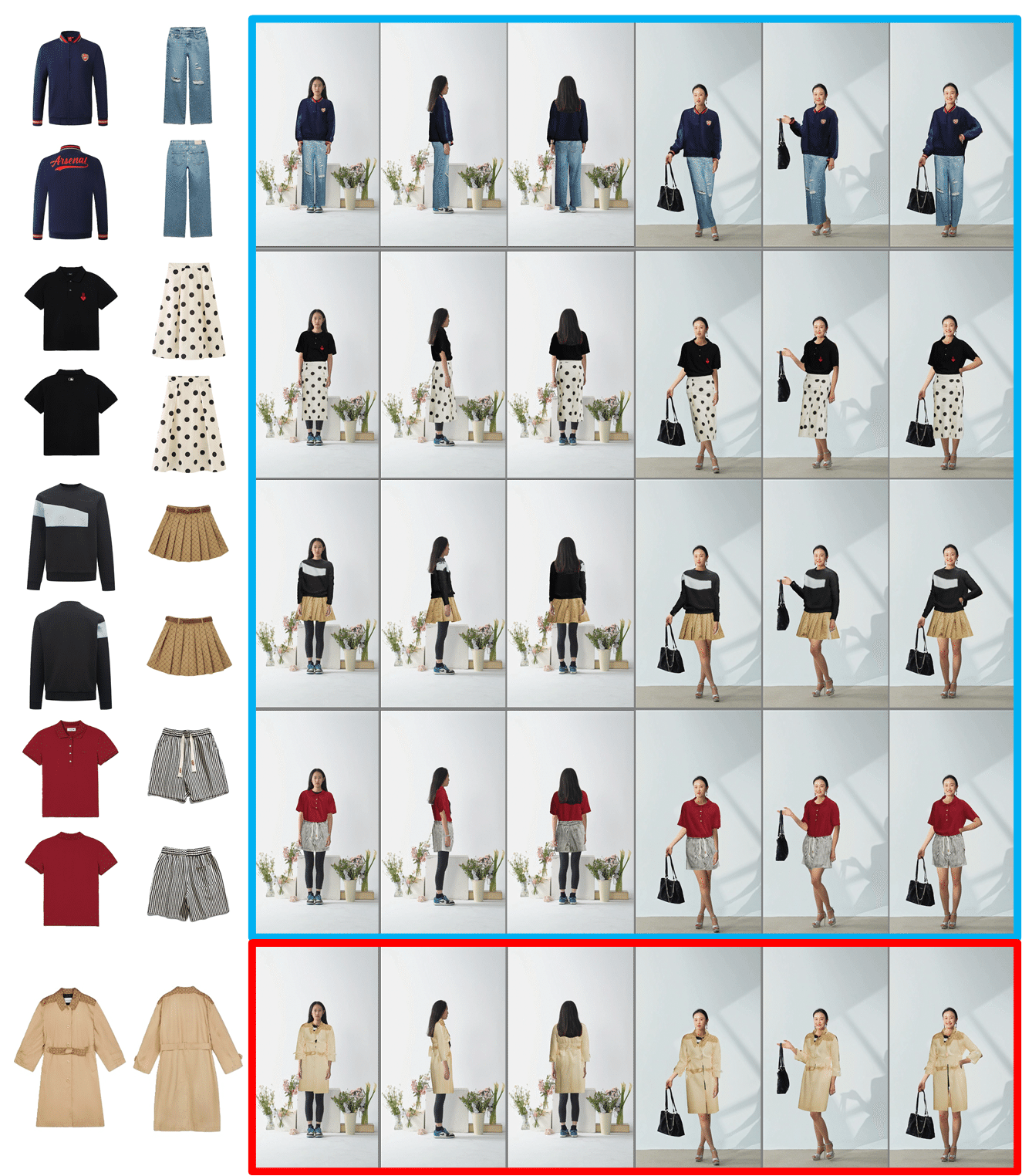}
  \caption{Visualization of 3D virtual try-on. We obtain textured 3D meshes from 2D reference images shown on the left. The 3D meshes are then draped onto 3D humans.}
\label{fig:sup_1}
\end{figure*}

\subsection{Limitations}

As shown in \cref{fig:sup_1}, \method can produce high-quality textures for common garments, \eg T-shirt, Shorts, Trousers and \etc ({\color{EpsilonColor}{blue}} bounding box (bbox)). However, we have observed that it is having difficulty in recovering textures for garments with complex patterns: \eg inaccurate and inconsistent local texture (belt, collarband) occurred in windbreaker ({\color{RedColor}{red}} bbox). We regard this as the extra accessories occurred in the garment, which inevitably add on the partial texture in addition to the main UV.

Another imperfection is that our method cannot maintain the uniformity of checked shirts with densely assembled grids: As shown in the second row of \cref{fig:qual_3}, our method inferior to 2D VTON methods in preserving texture among which comprised of thousands of fine and tiny checkerboard-like grids, checked shirts and pleated skirts are representative type of such garments.

We boil this down to the subtle position changes during deformation graph optimization period, which leads to the template mesh becomes less uniform eventually as the regularization terms, \ie as-rigid-as-possible is not a very strong constraint energy terms in obtaining a conformal mesh. We acknowledge this challenge and leave it to future work to explore the possibility in generating a homogeneous mesh with uniformly-spaced triangles.
\section{Conclusion}

This paper presents a novel pipeline, \method, for synthesizing high-quality textures for 3D meshes from the pictures taken from only front and back views. \method adopts a two-stage process in obtaining visually appealing textures, where phase \RNum{1} offers coarse texture generation and phase \RNum{2} performs texture refinement. Training a generalized texture inpainting network is non-trivial due to the high topological variability of UV space. Therefore, obtaining paired data under such circumstances is important. To the best of our knowledge, this is the first study to combine a diffusion model with a 3D engine (Blender) in collecting coarse-fine paired textures in 3D texturing tasks. We show the generalizability of this approach in a variety of examples. 

To avoid distortion and stretched artifacts across clothes, we automatically adjust the scale of vertices of template meshes and thus best prepare them for later image-based optimization, which effectively guides the implicitly learned texture with a complete and distortion-free structure. Extensive experiments demonstrate that our method can effectively synthesize consistent and highly detailed textures for typical clothes without extra manual effort. 

In summary, we hope our work can inspire more future research in 3D texture synthesis and shed some light on this area.

\hspace*{\fill}
{
    \bibliographystyle{plainnat}
    \bibliography{main}
}

\setcounter{page}{1}
\maketitlesupplementary

\section{Implementation Details}
In phase \RNum{1}, we fix the optimization steps of both silhouette matching and image-based optimization to 1,000, which makes each coarse texture generation process takes less than 1 minute to complete on an NVIDIA Ampere A100 (80GB VRAM). The initial weights of each energy term are $w_{sil} = 50, w_{lmk} =0.01, w_{arap} =50, w_{norm}=10, w_{img}=100, w_{tv}=1$, we then use cosine scheduler for decaying $w_{arap}, w_{norm}$ to $5, 1$.

During the blender-enhanced rendering process, we augment the data by random sampling blendshapes of upper cloth by a range of $\left[0.1, 1.0\right]$. The synthetic images were rendered using Blender \textbf{EEVEE} engine at a resolution of $512^{2}$, emission only (disentangle from the impact of shading, which is the notoriously difficult puzzle as dissected in Text2Tex~\cite{chen2023text2tex}). 

\begin{figure*}[htpb]
\includegraphics[width=1.0\linewidth]{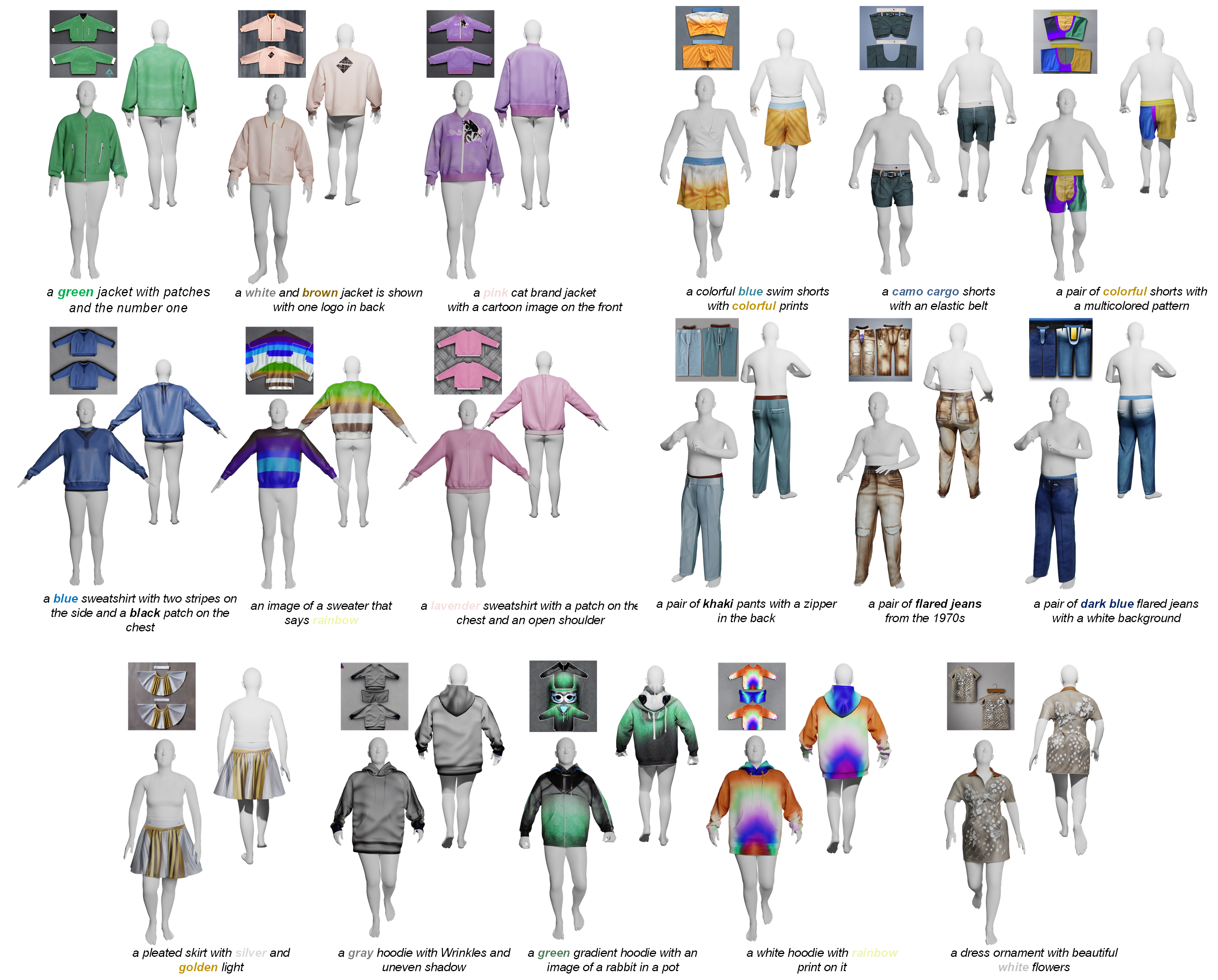}
  \caption{Texture maps for training instance map guided Pix2PixHD, synthesized by ControlNet \textit{canny edge}.}
\label{fig:sup_2}
\end{figure*}

The synthetic data used for training texture inpainting network are yielded from pretrained ControlNet through prompts (generates from Lavis-BLIP~\cite{lavis}, OFA~\cite{ofa} and MPlug~\cite{mplug}) and UV templates (manually crafted UV maps by artists) can be shown in \cref{fig:sup_2}, which contains more garment types than previous methods, \eg Pix2Surf~\cite{pix2surf} (\textbf{4}) and Warping~\cite{majithia2022robust} (\textbf{2}).

The only existing trainable Pix2PixHD in phase \RNum{2} is optimized by Adam~\cite{kingma2014adam} with $lr=2e-4$ for $200$ epochs. Our implementation is build on top of PyTorch~\cite{pytorch} alongside PyTorch3D~\cite{ravi2020pytorch3d} for silhouette matching, rendering and inpainting.

\begin{table}[h]
\vspace{-0.5em}
\caption{SOTA inpainting methods act on our synthetic data.}
\label{tab:sup1}
\centering
\begin{tabular}{@{}ccc@{}}
\toprule
 Baseline & Inpainting & SSIM $\uparrow$ \\
\hline
Phase \RNum{1}  & \textit{None}  & 0.80  \\
Phase \RNum{1}  & \textit{Navier-Stokes}~\cite{nsinpainting}  &  0.80     \\
Phase \RNum{1}  & \textit{LaMa}~\cite{lama}  &  0.78    \\
Phase \RNum{1}  & \textit{Stable Diffusion (v2)}~\cite{stablediffusion}  &  0.77    \\
Phase \RNum{1}  & \textit{Deep Floyd}~\cite{DeepFloyd}  &  0.80    \\
\bottomrule
\end{tabular}
\end{table}

\begin{table}[h]
\vspace{-0.5em}
\caption{Inpainting methods trained on our synthetic data.}
\label{tab:sup2}
\centering
\begin{tabular}{@{}ccc@{}}
\toprule
 Baseline & Inpainting & SSIM $\uparrow$ \\
\hline
Phase \RNum{1}  & \textit{None}  & 0.80  \\
Phase \RNum{1}  & \textit{Cond-TransUNet}~\cite{chen2021transunet}   & 0.78  \\
Phase \RNum{1}  & \textit{ControlNet}~\cite{controlnet}   & 0.77  \\
Phase \RNum{1}  & \textit{Pix2PixHD}~\cite{pix2pixhd}   & \textbf{0.83} \\
\bottomrule
\end{tabular}
\end{table}

\begin{figure}[htpb]
\includegraphics[width=1\linewidth]{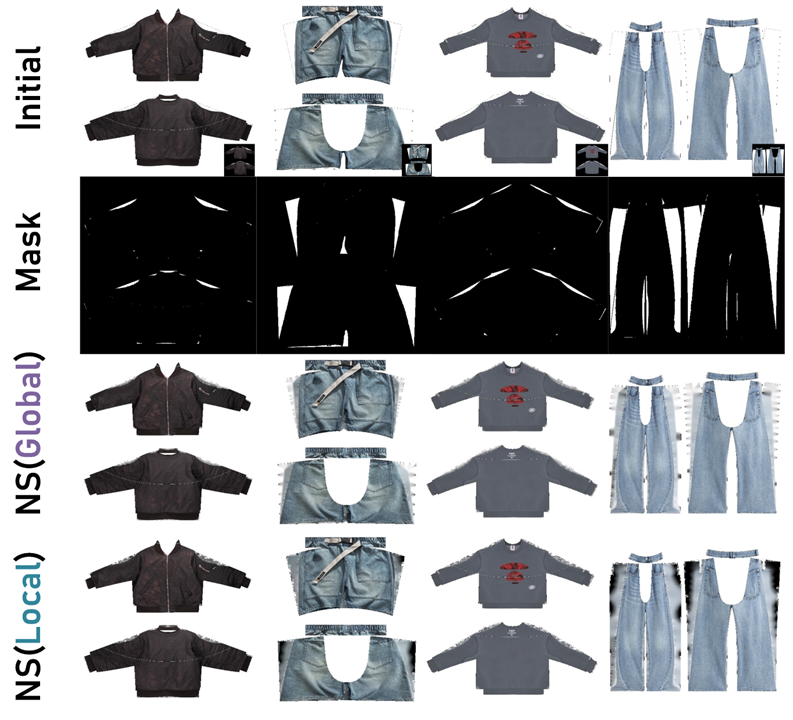}
  \caption{Visualization of Navier-stokes method on UV template. Our locally constrained NS method fills the blanks thoroughly (though lack of precision) compared to the original global counterpart.}
\label{fig:sup1}
\end{figure}

\begin{figure*}[htpb]
\includegraphics[width=1.0\linewidth]{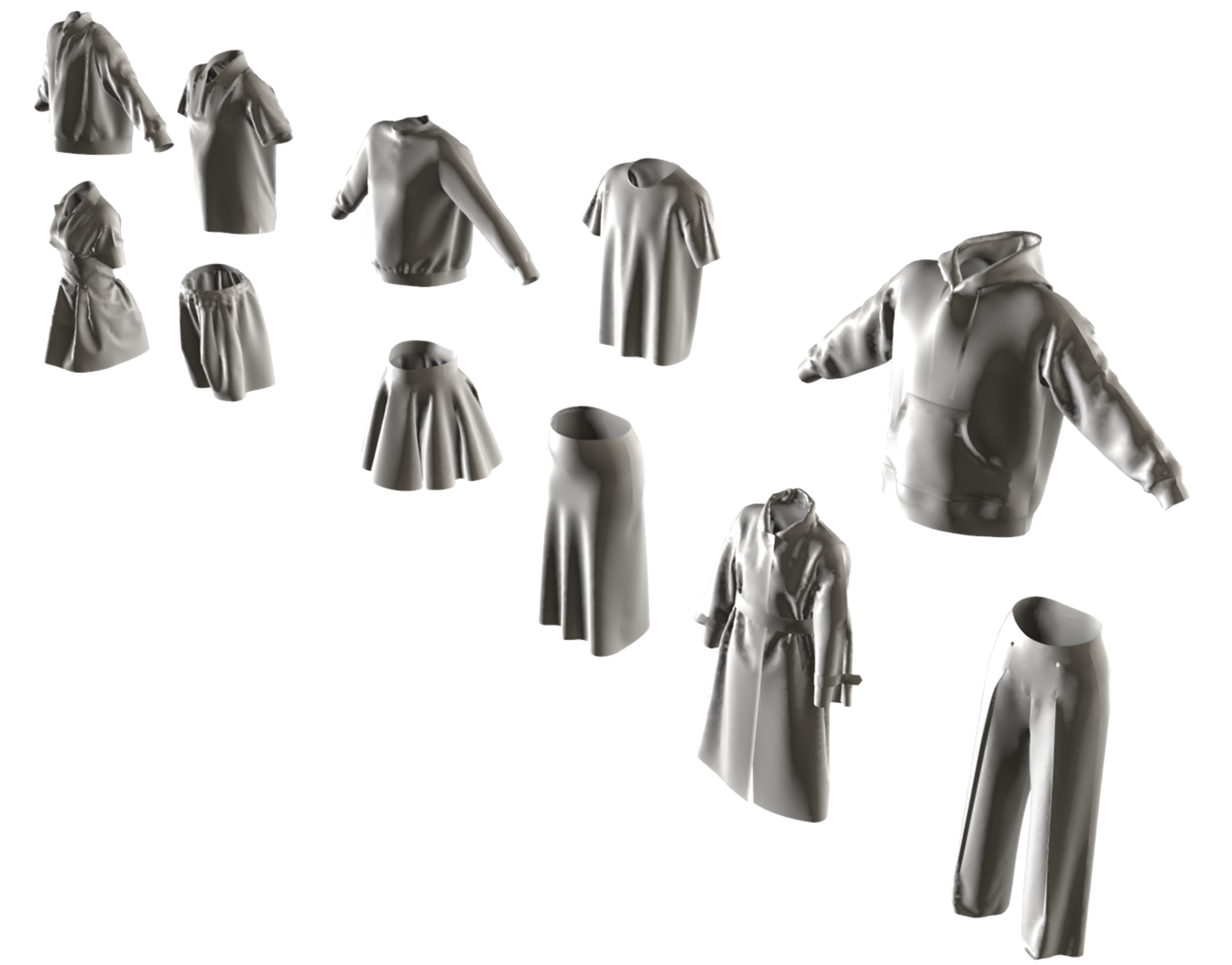}
  \caption{Visualization of all template meshes used in \method.}
\label{fig:sup0}
\end{figure*}

\begin{figure*}[htpb]
\includegraphics[width=1.0\linewidth]{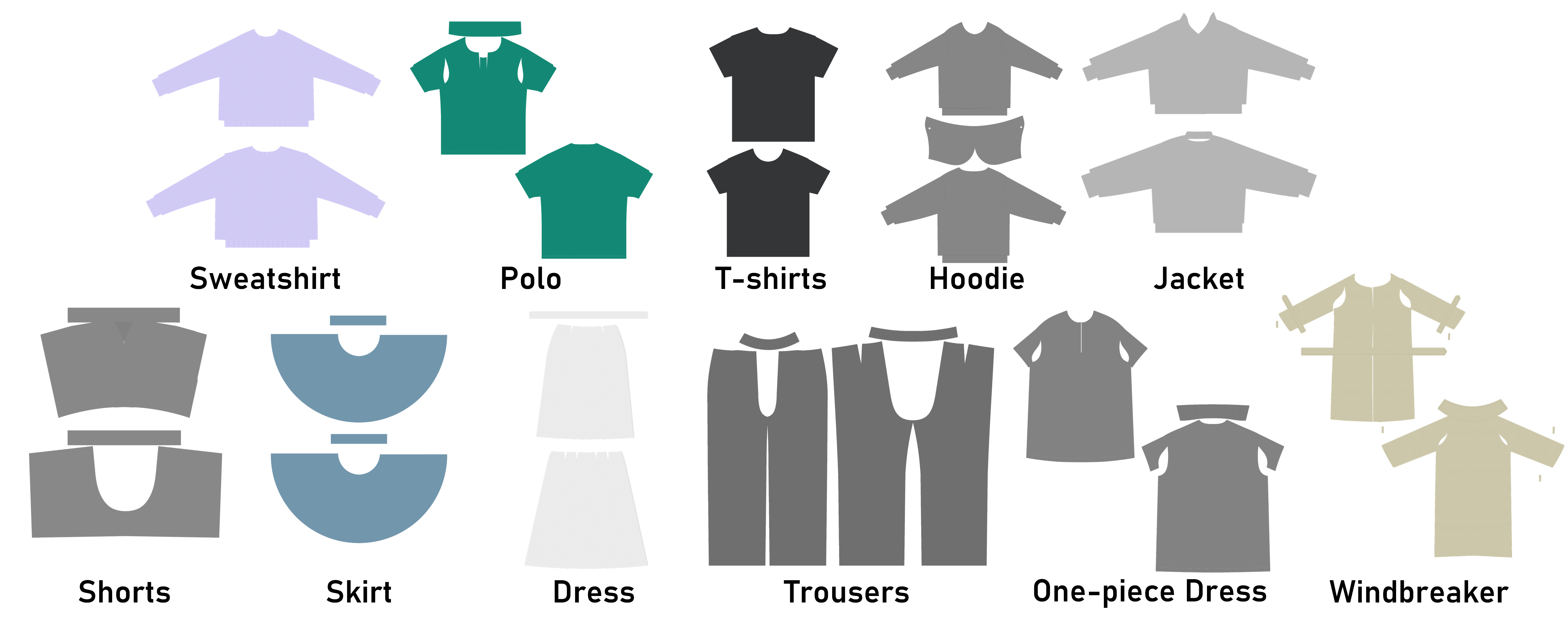}
  \caption{All UV maps of template meshes used in \method.}
\label{fig:sup_tex}
\end{figure*}

The detailed parameters of template meshes in \method are summarized in \cref{tab:mesh_comp}, sketch of all template meshes and UV maps are shown in \cref{fig:sup0} and \cref{fig:sup_tex} respectively.

\begin{table*}[htpb]
\caption{Detailed parameters of template mesh in \method. As shown in the table, each template's vertex is less than 10,000 and all are animatable by means of Style3D, which is the best fit software for clothing animation.}
\label{tab:mesh_comp}
\centering
\begin{tabular}{l|c|c|c|c}
\toprule
Category & Vertices & Faces &  Key Nodes (Deformation Graph)  &  Animatable \\
\hline
T-shirts  & 8,523    & 16,039    & 427   & \cmark \\
Polo      & 8,922    & 16,968    & 447    & \cmark \\
Shorts       & 8,767    & 14,845   & 435    & \cmark \\
Trousers      & 9,323    & 16,995   & 466    & \cmark \\
Dress & 7,752    & 14,959    & 388    & \cmark \\
Skirt & 6,116    & 11,764    & 306    & \cmark \\
Windbreaker & 9,881    & 17,341    & 494    & \cmark \\
Jacket & 8,168    & 15,184    & 409    & \cmark \\
Hoodie (Zipup) & 8,537    & 15,874    & 427    & \cmark \\
Sweatshirt & 9,648    & 18,209    & 483    & \cmark \\
One-piece Dress & 9,102    & 17,111    & 455    & \cmark \\
\bottomrule
\end{tabular}
\end{table*}

\begin{figure*}[htpb]
\includegraphics[width=1.0\linewidth]{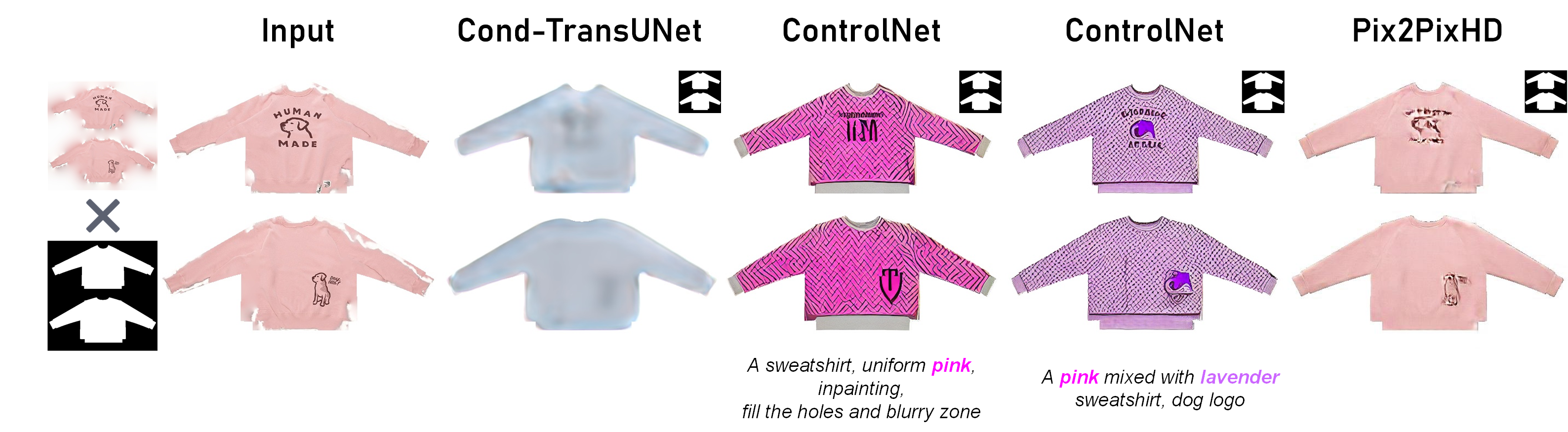}
  \caption{Comparison with representative image2image methods with conditional input: autoencoder-based TransUNet~\cite{chen2021transunet} (we modify the base model and add an extra branch for UV map, aims to train it with all types of garment together), diffusion-based ControlNet~\cite{controlnet} and GAN-based Pix2PixHD~\cite{pix2pixhd}. It is rather obvious that prompts-sensitive ControlNet limited in recover a globally color-consistent texture maps. Upper right corner of each method is the conditional input.}
\label{fig:sup_3}
\end{figure*}

\section{Self-modified UV-constrained Naiver-Stokes Method}

As shown in \cref{fig:sup1}, we display the results between our self-modified UV-constrained Navier-Stokes (NS) method (\textit{local}) and original NS (\textit{global}) method. Specifically, we add a reference branch (UV template) for NS and thus confine the inpainting-affected region to the given UV template for each garment, thus contributing directly to the interpolation result. Our locally constrained NS method allows blanks to be filled thoroughly compared to the original global NS method.

The sole aim of modifying the original global NS method is to conduct a fair comparison with deep learning based methods as depicted in the main paper.

The noteworthy thing is that for small blank areas (\eg Column \textit{1,3} of \cref{fig:sup1}), the texture uniformity and consistency are well-persevered thus capable of producing plausible textures.

\section{Efficiency of mainstream Inpainting methods}

As depicted in the main paper, our neural rendering based pipeline achieves superior SSIM compared to TPS warping. This improvement is also preserved after inpainting and refinement, leading to a much better quality of the final texture. 

Free from the page limit in the main paper, here we conduct a comprehensive comparison study on various inpainting methods act upon the coarse texture maps derived from Phase \RNum{1} directly, to demonstrate the efficiency of mainstream inpainting methods.

First, we compare the state-of-the-art inpainting methods quantitatively on our synthetic coarse-fine paired dataset. One thing to note is that checkpoints derived from all deep learning based inpainting methods are open and free. No finetune or modification is involved in this comparison.
As described in \cref{tab:sup1}, none of such methods produce a noticeable positive impact in boosting the SSIM score compared to the original coarse texture (\textit{None} version).

Next, we revise TransUNet~\cite{chen2021transunet} with input a conditional UV map for the unity of the input and output with ControlNet~\cite{controlnet} and Pix2PixHD~\cite{pix2pixhd}. Then we train \textit{cond-TransUNet}, ControlNet, and Pix2PixHD on the synthetic data for a fair comparison. We input all these three with original input coarse texture maps, conditional input UV maps, and output fine texture maps. The selective basis of TransUNet, ControlNet, and Pix2PixHD originates from the generative paradigm: TransUNet is a basic autoencoder-based supervised learning image2image model, ControlNet is a diffusion-based generative model and Pix2PixHD is a GAN-based generative model. We want to explore the feasibility of these methods in our task, as depicted in \cref{tab:sup2} and \cref{fig:sup_3}, Pix2PixHD is superior in obtaining satisfactory texture maps in terms of both qualitative and quantitative views.



\end{document}